\documentclass[runningheads]{llncs}

\usepackage[year=2026]{eccv}

\usepackage{eccvabbrv}   
\usepackage{graphicx}
\graphicspath{{figures/}}
\usepackage{booktabs}
\usepackage{amsmath}
\usepackage[accsupp]{axessibility}  

\usepackage{hyperref}
\usepackage{orcidlink}
\usepackage[capitalize]{cleveref}  

\newcommand{\ourmethod}{RoR}

\begin{document}

\title{Reason or Recite: Complementary Specialists with an LLM Judge for Traffic Anomaly Reasoning}

\titlerunning{Reason or Recite}

\author{Mayur Kulkarni}

\authorrunning{M.~Kulkarni}

\institute{Stellarview AI, Pune, India \\
\email{mayur.k@stellarview.ai}}

\maketitle

\begin{abstract}
Road anomalies such as collisions, violations, and sudden hazards demand
more from video understanding than detection: a practical system must verify
what happened, describe the scene, and explain the cause. We find these
competences pull in opposite directions for vision-language models:
closed-form questions reward precise, evidence-grounded reasoning, while
open-ended answers are scored by similarity to reference text, where matching
the annotation style matters as much as reasoning correctly. We present
Reason-or-Recite (\ourmethod{}), a two-specialist framework for the Traffic
Anomaly Reasoning track of the 2026 AI City Challenge. \ourmethod{}
(1) rewrites the official chain-of-thought annotations into a structured,
temporally grounded schema and LoRA-tunes a 9B vision-language model on the
enhanced traces; (2) fine-tunes a 4-bit-quantized 27B model on answers alone,
absorbing the annotation style; and (3) selects between the two
candidates per question with a training-free LLM judge. \ourmethod{} ranks
first on the in-domain leaderboard with mean score 0.6788.
\end{abstract}

\section{Introduction}
\label{sec:intro}

Understanding anomalous events in transportation goes well beyond flagging
that something unusual happened: a practical system must verify whether a
specific event occurred, describe the scene and actors, localize the
event in time, and explain \emph{why} it happened. Track~3 of the 2026 AI
City Challenge~\cite{Tang26AICity26}
operationalizes this via the Traffic Anomaly Reasoning (TAR) benchmark: a
unified video-language model must answer ten heterogeneous task types, nine
of them scored, over
CCTV traffic footage, ranging from binary event verification and
multiple-choice QA to open-ended causal analysis, scene description, and video
summarization.

Two properties of the benchmark shape our solution. First, the task types
demand \emph{different competences}: closed-form questions (binary and
multiple-choice) reward precise, evidence-grounded reasoning, while the seven
open-ended tasks are scored with BERTScore~\cite{zhang2020bertscore} against
a single reference answer. Second, similarity to a single reference rewards
matching that reference's register and content-selection policy as much as
answering correctly, and the training annotations, machine-generated by a VLM
auto-labeling pipeline (\cref{sec:task-annotations}), are the only available
sample of the style being scored against: a model that reasons \emph{better}
than its supervision can paradoxically score \emph{worse}.

We therefore decouple the two competences. For evidence-grounded reasoning, we
rewrite the official chain-of-thought annotations into a structured,
evidence-grounded schema
(\texttt{<scene>}, \texttt{<query>}, \texttt{<evidence>} with timestamps,
\texttt{<reason>}, \texttt{<conclusion>}) using an off-the-shelf LLM, and
LoRA-tune~\cite{hu2022lora} a Qwen~3.5~9B VLM on the enhanced traces
(\cref{sec:method-tarre,sec:method-reasoner}). For open-ended tasks, we
fine-tune a larger Qwen~3.6~27B model with QLoRA~\cite{dettmers2023qlora} on
the \emph{answers only}, letting it absorb the training annotation
distribution directly
(\cref{sec:method-answerer}). At inference, a frozen Qwen~3.6~27B
\emph{judge} selects the better candidate per question, with no
additional training (\cref{sec:method-judge}).

Our contributions are:
\begin{itemize}
  \item \textbf{Structured reasoning enhancement (TAR-RE):} an LLM-driven
  re-annotation of the TAR reasoning traces into an explicit, temporally
  grounded schema, substantially improving closed-form QA after
  fine-tuning.
  \item \textbf{Answer-distribution specialization for reference-scored
  generation:} an answer-only QLoRA fine-tune of a larger model that
  absorbs the auto-labeled annotation distribution directly and
  outperforms reasoning-tuned models on the BERTScore-scored open-ended
  tasks, showing that under single-reference scoring, matching the
  annotation distribution can outweigh gains in reasoning quality.
  \item \textbf{Evidence-gated per-question arbitration:} a training-free
  LLM judge that selects between the specialists per question;
  unconstrained, it scores below the stronger specialist due to
  position and family bias, while our conservative design gains over
  both.
  \item \textbf{First place:} \textbf{1st} on the Track~3
  in-domain (TAR) leaderboard with mean score \textbf{0.6788}, including
  perfect binary event verification.
\end{itemize}

\section{Related Work}
\label{sec:related}

\subsection{Vision-Language Models for Video Understanding}
Open multimodal LLMs have matured rapidly into capable video
reasoners~\cite{bai2023qwenvl,wang2024qwen2vl,bai2025qwen25vl,lin2024vila,zhang2024llavavideo,zhu2025internvl3},
with recent backbones built for long, temporally rich
video~\cite{wang2025internvideo25}: Qwen3-VL interleaves textual
timestamps with video tokens over a 256K context~\cite{bai2025qwen3vl}. On the
language side, Qwen3~\cite{yang2025qwen3} integrates a fast non-thinking
mode and an explicit thinking mode into a single model, a
reasoning/answering duality our two-specialist design mirrors at the system
level. We build on this line: the reasoning specialist applies
LoRA~\cite{hu2022lora} to Qwen~3.5~9B~\cite{qwen2026qwen35} and the answer
specialist QLoRA~\cite{dettmers2023qlora} to
Qwen~3.6~27B~\cite{qwen2026qwen36}, its architecturally identical, further
post-trained successor.
A complementary line grounds video reasoning in time, via timestamp-bound
features, supervised event boundaries, timestamped causal traces, or
temporally indexed evidence
banks~\cite{ren2024timechat,huang2024vtimellm,guo2025trace,wang2025videotree,fan2024videoagent};
rather than encoding time in the architecture or the retrieval stage, our
structured traces make timestamped evidence explicit in the supervision
itself.

\subsection{From Anomaly Detection to Reasoning Supervision}
Traffic and surveillance anomaly analysis began as \emph{detection}: weakly
supervised methods score video snippets from video-level
labels~\cite{sultani2018real,tian2021weakly}, supported by traffic-specific
benchmarks spanning dashcam and CCTV
views~\cite{fang2019dada,yao2022dota,lv2021tad,xu2024tad,chen2024sotad}.
Such systems localize anomalies but cannot explain them, so a second wave
reframes anomaly analysis as \emph{language understanding} via multimodal
LLMs~\cite{du2024uncovering,tang2024hawk,zhang2024holmesvad}, prompted
foundation models~\cite{zanella2024harnessing}, or verbalized
learning~\cite{ye2025vera}; most recently, VAD-R1~\cite{huang2025vadr1} and
VAU-R1~\cite{zhu2025vaur1} make the model \emph{reason} before answering,
optimizing the reasoning with GRPO-style reinforcement learning.
How that reasoning should be supervised is a question of its own:
chain-of-thought prompting improves answer accuracy~\cite{wei2022chain},
the \emph{structure} of the trace matters~\cite{chen2023pot,yao2023tot},
distillation trains directly on
rationales~\cite{hsieh2023distilling,zelikman2022star,mukherjee2023orca},
with recent evidence that \emph{rewriting} traces to fit the student is what
makes it succeed~\cite{kim2025rsd}, while reinforcement learning abandons
trace supervision altogether~\cite{deepseekai2025r1,feng2025videor1}.
Auto-generated supervision, finally, carries the generator's stylistic
fingerprint~\cite{panickssery2024selfpreference}, an effect central to our
setting (\cref{sec:task-annotations}). We improve the \emph{supervision}
rather than the optimizer: restructuring weak auto-labeled traces into an
evidence-grounded schema is far cheaper than RL rollouts and remains
compatible with lightweight LoRA fine-tuning.

\subsection{Output Selection and LLM-as-Judge}
Reliability can be improved by choosing among candidate outputs rather than
trusting a single decode: self-consistency marginalizes over sampled
reasoning paths~\cite{wang2023selfconsistency}, LLM-Blender ranks and fuses
outputs from multiple models~\cite{jiang2023llmblender}, strong LLMs have
proven reliable automated judges of response
quality~\cite{zheng2023judging}, and routers such as
RouteLLM~\cite{ong2024routellm} instead pick a model \emph{before} inference
to trade off cost and quality. Our judge differs on both axes:
training-free, it arbitrates \emph{after} generation between two
heterogeneous fine-tuned specialists, per question.

\subsection{VLM Solutions in the AI City Challenge}
Successive AI City Challenges~\cite{wang2024eighth,tang2025ninth} have
pushed VLMs into fine-grained traffic understanding:
CityLLaVA~\cite{duan2024cityllava} established bounding-box visual prompting
and best-view selection, later extended with domain-aware expert priors and
QLoRA tuning~\cite{ha2025domainaware}; other strong entries decompose
captions into spatial and temporal segments~\cite{nguyennhu2025stervlm} or
combine spatially guided fine-tuning with caption
refinement~\cite{phimsiri2025trafficinternvl}. Closest to our design,
Kyem~\etal~\cite{kyem2025taskspecific} route statically by task type
between two specialists, and Kachhadiya~\etal~\cite{kachhadiya2025multiagent}
dispatch each question type to the agent with the best validation accuracy;
both rely on a precomputed routing table, whereas our judge
arbitrates per question; our annotation enhancement restructures
reasoning traces rather than decomposing captions.

\section{Task and Dataset}
\label{sec:task}

\subsection{The TAR Benchmark}
\label{sec:task-benchmark}
The TAR training set comprises 44{,}040 annotations over 3{,}670 CCTV videos
(965 anomalous, 2{,}705 normal; ${\sim}26.1$ hours) drawn from eight public
datasets~\cite{Tang26AICity26}, including
UCF-Crime~\cite{sultani2018real}, TAD~\cite{lv2021tad,xu2024tad},
SO-TAD~\cite{chen2024sotad}, VAD-R1~\cite{huang2025vadr1}, and
Accident-Bench~\cite{gu2025accidentbench}.
Each annotation attaches a question, its gold answer, and a chain-of-thought
reasoning trace to a video.
The ten task types fall into three groups: \emph{Basic} (event
verification (binary), event verification with explanation,
multiple-choice QA, multiple-choice QA with explanation, and open-ended
QA); \emph{Scene} (scene description and video summarization); and
\emph{Temporal} (temporal localization, causal linkage, and event
description). The in-domain test set contains 960 human-verified annotations
over 80 clips trimmed from 17 public YouTube videos.
Binary and multiple-choice answers are scored by exact accuracy (regex
extraction), all open-ended tasks by BERTScore
F1~\cite{zhang2020bertscore} (roberta-large, rescaled); the leaderboard
mean is the unweighted average over nine task types (temporal localization
was excluded from scoring by the organizers).

\subsection{How the Annotations Were Made, and Why It Matters}
\label{sec:task-annotations}

\begin{figure}[t]
  \centering
  \includegraphics[width=\textwidth]{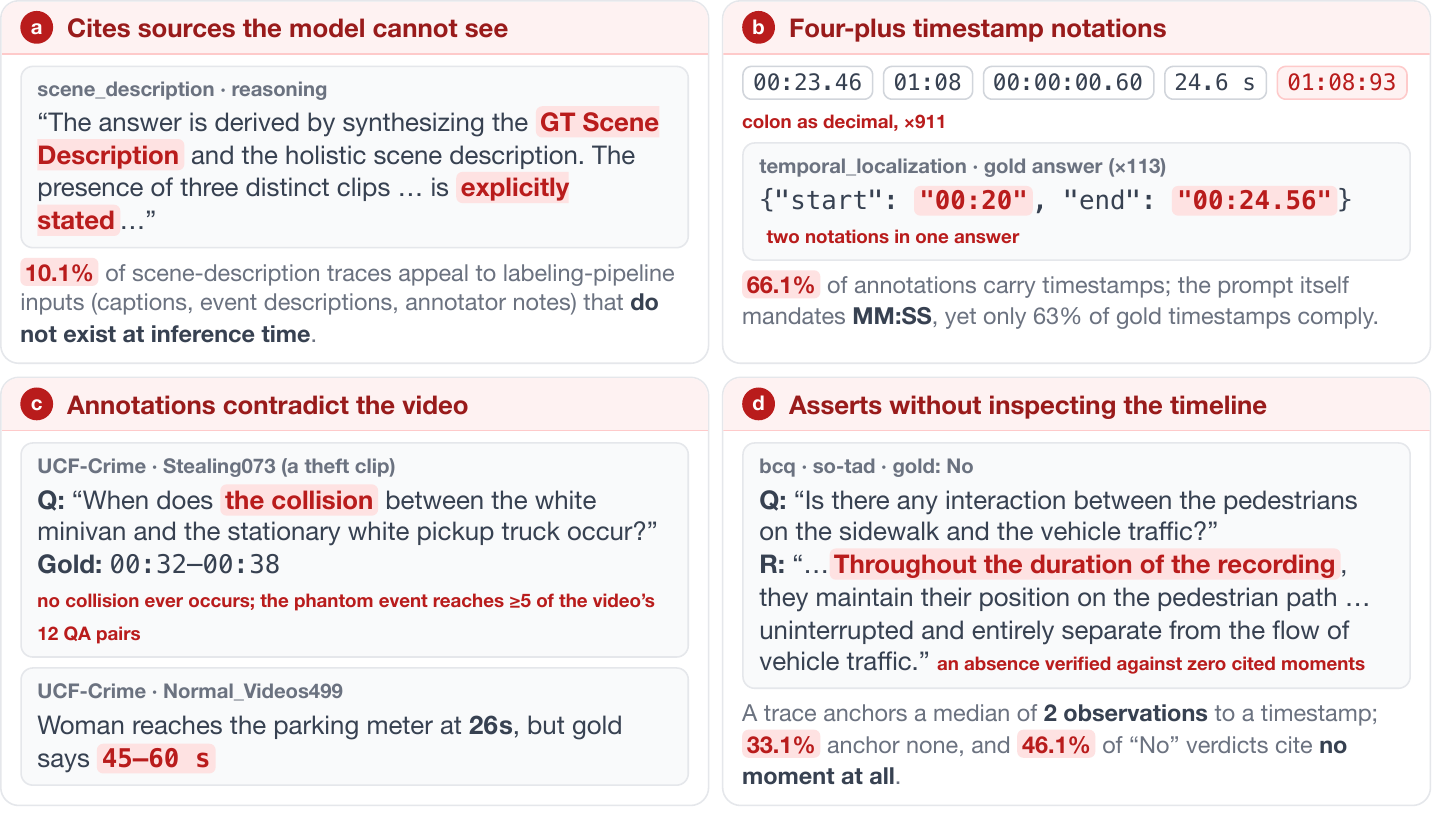}
  \caption{Four defect classes of the official TAR annotations, with
  verbatim training examples; each is detailed in
  \cref{sec:task-annotations}. Panel~(c) shows
  UCF-Crime~\cite{sultani2018real} clips.}
  \label{fig:defects}
\end{figure}

The 44{,}040 training annotations are machine-generated by a three-stage
VLM auto-labeling pipeline~\cite{Tang26AICity26}: Gemini 3.1 Pro produces
a caption and a structured event description for each video, and Gemma-4
converts these into multi-task Q\&A pairs with chain-of-thought reasoning.
For 910 of the 3{,}670 videos, pre-existing NVIDIA human annotations
served as supplementary context; the test references, in contrast, are
human-verified~\cite{Tang26AICity26}.
Before training anything, we examined the training annotations, with
programmatic checks across the full set and manual inspection of a small
sample of videos; the four defect classes this surfaced
(\cref{fig:defects}, quantified in \cref{tab:audit}) shaped every
design decision in \ourmethod{}.

\paragraph{(a) The reasoning cites sources the model cannot see.}
Reasoning distilled from a labeling pipeline inherits the pipeline's inputs:
many traces justify an answer not from the video but from the captions and
annotator notes the teacher models were prompted with (\cref{fig:defects}a).
Conservative phrase matching (``ground truth'', ``the provided
description'', \etc) flags 10.1\% of scene-description reasoning and
${\sim}1\%$ overall, a lower bound since only explicit wording is
detectable; fine-tuning on such traces teaches a model to appeal to an
authority it will never have.

\paragraph{(b) Timestamps are everywhere, in no consistent format.}
Temporal grounding permeates the supervision: 66.1\% of all annotations
contain at least one timestamp, most inside the reasoning itself
(\cref{tab:audit}), in chaotic notation: four conventions coexist, often
within a single item, 911 timestamps use a colon as a decimal separator,
and 113 gold answers mix two conventions inside one JSON object
(\cref{fig:defects}b).

\paragraph{(c) Some annotations contradict the video.}
Manual inspection of sampled clips surfaced annotations whose premise or
timing is simply wrong (\cref{fig:defects}c), including an entire question
family premised on a collision that never occurs: all task types derive
from the same event description, so one upstream hallucination propagates
into every task for that video. We deliberately do not relabel, since gold
outcomes anchor our regeneration (\cref{sec:method-tarre}), but such noise
bounds what any method can extract from this training distribution.

\paragraph{(d) The reasoning asserts without inspecting the timeline.}
Excluding scene description, whose global character needs no timestamps,
a trace anchors a median of two observations to a timestamp, and 33.1\%
anchor none. The verification case is sharpest: when
a binary question's gold answer is \emph{No}, establishing that absence
requires scanning the whole video, yet 46.1\% of ``No'' traces cite no
timestamp (\cref{fig:defects}d), teaching a model to output verdicts about
a timeline it never inspects.

Beyond these four classes, the traces are uniformly shallow: the mean
trace length, computed per task type, ranges from 109 words on the shortest
task to 244 on the longest, and the traces are largely a restatement of the answer they
should derive (token-level ROUGE-L~\cite{lin2004rouge} between the two
reaches 0.33; \cref{tab:audit}). Defects (a)--(d) and this shallowness
motivate regenerating the reasoning specialist's supervision
(\cref{sec:method-tarre}). The same provenance cuts the other way, too:
these machine-generated answers are the only available sample of the
register the open-ended tasks are scored against, the property the
answer specialist exploits (\cref{sec:method-answerer}).

\begin{table}[t]
  \centering
  \caption{Auditing the official TAR training annotations, and the effect of
  reasoning regeneration (TAR-RE, \cref{sec:method-tarre}). TS = share of
  annotations containing at least one timestamp; Words = mean reasoning
  length; R-L = mean token-level ROUGE-L F1 between a reasoning trace and
  its gold answer (higher = more restatement; omitted where the answer is a
  single word, option letter, or timestamp pair). Plain BCQ and MCQ match
  their -OE rows to within 0.3 and are omitted.}
  \label{tab:audit}
  \setlength{\tabcolsep}{5pt}
  \begin{tabular}{lrr rr rr}
    \toprule
    & & & \multicolumn{2}{c}{Words} & \multicolumn{2}{c}{R-L vs.\ answer} \\
    \cmidrule(lr){4-5}\cmidrule(lr){6-7}
    Task & \#Items & TS\,\% & Official & TAR-RE & Official & TAR-RE \\
    \midrule
    BCQ-OE               & 7{,}340 &  59.8 & 109 & 429 & 0.128 & 0.058 \\
    MCQ-OE               & 3{,}670 &  53.0 & 123 & 463 & 0.141 & 0.065 \\
    Open QA              & 3{,}670 &  57.2 & 126 & 471 & 0.246 & 0.151 \\
    Scene description    & 3{,}670 &   3.0 & 118 & 483 & 0.326 & 0.212 \\
    Video summarization  & 3{,}670 &  89.0 & 117 & 482 & 0.289 & 0.233 \\
    Temporal localization& 3{,}670 & 100.0 & 139 & 468 & --    & --    \\
    Event description    & 3{,}670 & 100.0 & 164 & 485 & 0.203 & 0.147 \\
    Causal linkage       & 3{,}670 & 100.0 & 244 & 505 & 0.276 & 0.195 \\
    \bottomrule
  \end{tabular}
\end{table}

\section{Method}
\label{sec:method}

\subsection{Overview}
\label{sec:method-overview}

\ourmethod{} turns the diagnosis of \cref{sec:task} into a system in three
parts: TAR-RE (\cref{sec:method-tarre}) repairs the training supervision;
two specialists are then fine-tuned, one per scoring regime: the
reasoning specialist on the regenerated TAR-RE traces
(\cref{sec:method-reasoner}), the answer specialist on the gold answers
alone, with the reasoning stripped entirely (\cref{sec:method-answerer});
and at inference both specialists answer every test question
independently, and a frozen, training-free judge selects between the two
candidates per question (\cref{sec:method-judge}).

\paragraph{MCQ with explanation is answered by construction.}
Each MCQ-OE test item restates an MCQ item on the same video: the same
stem and candidate options, re-shuffled and occasionally lightly
re-worded. The multiple-choice inference therefore already contains the
open-ended answer, and we generate nothing for MCQ-OE: the option text
of the final MCQ prediction is matched to the corresponding option of
the MCQ-OE twin, and that option's wording is submitted verbatim: a
benchmark-specific shortcut, orthogonal to the method.
Under single-reference BERTScore (\cref{sec:task-annotations}),
reproducing the correct option's exact wording is the strongest
submission available whenever the underlying MCQ prediction is right.

\subsection{Structured Reasoning Enhancement (TAR-RE)}
\label{sec:method-tarre}

TAR-RE (\emph{TAR Reasoning Enhanced}) repairs the supervision defects of
\cref{sec:task-annotations} in two steps: normalize time, then regenerate the
reasoning.

\begin{figure}[t]
  \centering
  \includegraphics[width=\textwidth]{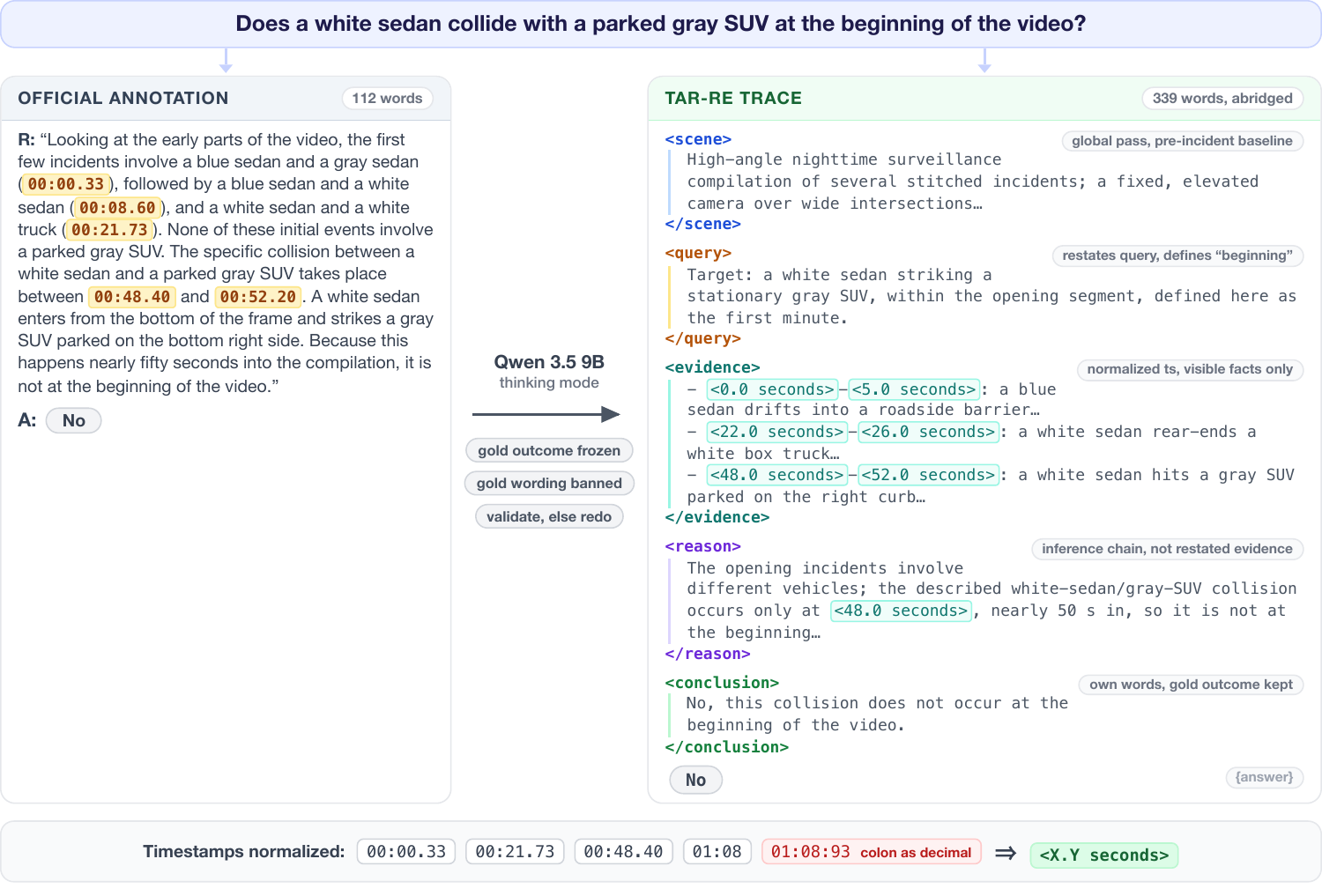}
  \caption{The TAR-RE rewrite (\cref{sec:method-tarre}) on one training
  item: the official annotation (left) is regenerated under our contract
  (center) into the five-part trace (right), preserving the gold outcome.
  \emph{Bottom:} timestamp canonicalization.}
  \label{fig:tarre-example}
\end{figure}

\paragraph{Unified timestamps.}
We first canonicalize every timestamp in every field to a single textual
form, \texttt{<X.Y seconds>}, parsing all the conventions catalogued in
\cref{sec:task-annotations} (\cref{fig:tarre-example}, bottom). Decimal
seconds need no base-60 arithmetic, and the format matches the textual
timestamps Qwen3-VL interleaves with video
tokens~\cite{bai2025qwen3vl}: the supervision speaks the backbone's
native representation of video time.

\paragraph{Regenerating the reasoning.}
We then use Qwen~3.5~9B in thinking mode to rewrite every training trace.
For each item, the model receives the video together with the task name, the
question, the gold answer, and the original reasoning; it deliberates in its
native thinking channel and must emit the five-part trace of
\cref{fig:tarre-example} (scene, query, timestamped evidence, reason,
conclusion), separating what is \emph{visible} (evidence) from what is
\emph{inferred} (reason), under the per-section depth floors shown there and
total floors of 220 words for closed-form and 320 words for descriptive
tasks: brevity, not verbosity, is the failure mode of the official traces.
Two constraints keep the rewrite honest: the gold \emph{outcome} is frozen
(the Yes/No polarity, option letter, or timestamp values must match the
official answer, so regeneration cannot silently relabel the data,
including, deliberately, the erroneous labels of
\cref{sec:task-annotations}), and the gold \emph{wording} is forbidden (a
validator rejects any trace that misses a structural tag, misses the
depth floors, or contains the gold answer verbatim; since the LLM samples
at non-zero temperature, rejected traces are simply regenerated until
they pass).

\paragraph{Why these five sections.}
The schema adapts the perception-to-cognition trace structure of
VAD-R1~\cite{huang2025vadr1} to the multi-task setting of TAR: global
before local, perception before cognition, visible fact separated from
inference. \texttt{<query>} pins down what to look for, since TAR spans
ten task types where VAD-R1 answers one fixed question;
\texttt{<evidence>} ties each observation to a timestamp; and
\texttt{<conclusion>} paraphrases the derived outcome, so the answer
that follows the think block restates a result rather than leaps to one.

\Cref{tab:audit} summarizes the effect: the traces become far longer yet
\emph{less} similar to their answers, with ROUGE-L overlap roughly halved
on every open-ended task. Over 99\% use the unified timestamp format
exclusively, the median number of time-anchored observations per trace
rises from two to 16--30, and the share of traces citing no moment at
all falls from 33.1\% to 0.2\% (scene description excluded). These
statistics measure structure, not factual grounding: defects (a), (b),
and (d) of \cref{sec:task-annotations} are removed by construction, but
nothing here certifies that every cited observation is correct; the
downstream effect on accuracy is measured in \cref{sec:exp-ablation}.

\paragraph{A joint target for the Scene-group tasks.}
Scene description and video summarization are the two tasks whose prompts
carry no distinguishing content. For the answer specialist
(\cref{sec:method-answerer}), rather than train them as separate items, we
merge each video's pair into a single coarse-to-fine target: the scene
description first, the summarization second, delimited as
\texttt{<scene>}\ldots\texttt{</scene>}\,\texttt{<summary>}\ldots\texttt{</summary>},
from which the two answers are sliced per task at inference; emitting the
global scene before the event narrative forces the model to commit to
coarse structure before fine-grained detail. The reasoning specialist, by
contrast, trains on all ten tasks in the single think-then-answer format
of \cref{sec:method-reasoner}.

\subsection{Reasoning Specialist: 9B LoRA on TAR-RE}
\label{sec:method-reasoner}

The reasoning specialist, a Qwen~3.5~9B VLM adapted to the TAR-RE traces
with LoRA~\cite{hu2022lora}, exists to \emph{reason before it answers}:
unlike the answer specialist, it is supervised to reproduce the full
trace, not merely the outcome.

\paragraph{Training target: think, then answer.}
We exploit Qwen~3.5's native thinking channel. For every task type, the
target places the complete five-part TAR-RE trace of
\cref{sec:method-tarre} inside the model's think block, and the gold outcome
immediately after it:
\[
  \texttt{<think>}\;\langle\text{scene}\rangle\ldots\langle\text{/conclusion}\rangle\;\texttt{</think>}\;\{\text{answer}\}.
\]
The question and video tokens are masked from the loss; supervision
covers trace and answer alike, and all ten task types share this one
format. The model deliberates in the channel the backbone was designed
for, and answer extraction is a single deterministic step: we decode
greedily, split on \texttt{</think>}, and read the text that follows, so
closed-form outputs reach the exact-match scorer without task-specific
parsing.

\paragraph{Adaptation.}
We fine-tune on the $80\%$ TAR-RE training split and keep the checkpoint
with the lowest validation loss; the adapter placement (shared with the
answer specialist), optimizer, schedule, and video sampling are in
\cref{sec:exp-impl}.

\begin{figure}[!t]
  \centering
  \includegraphics[width=\textwidth]{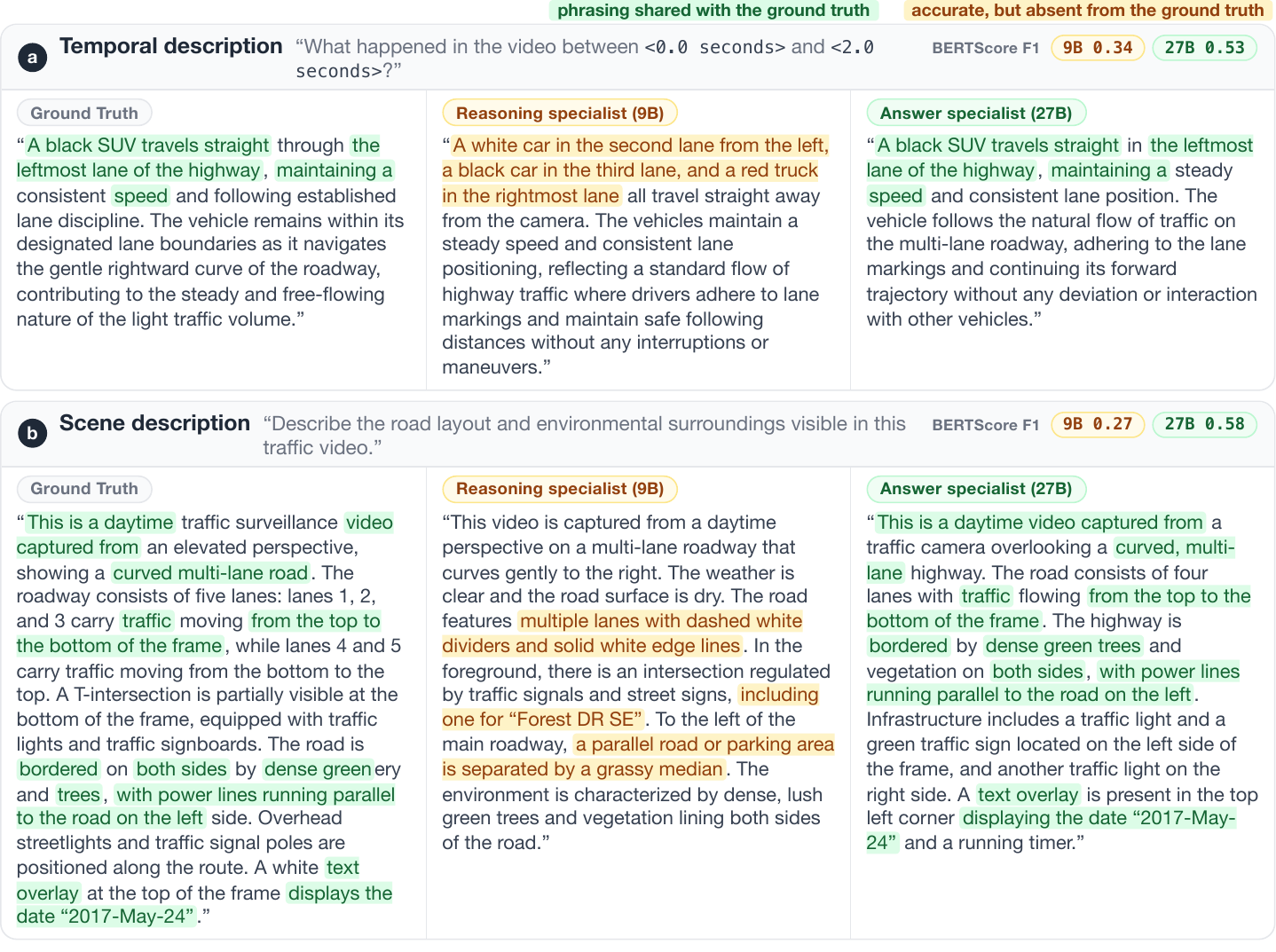}
  \caption{Reason or recite, on two open-ended validation items: the
  auto-labeled reference beside both specialists' answers; green marks
  phrasing shared with the reference, amber accurate but unrewarded
  content; BERTScore~F1 at top right. \emph{(a)}~The reference narrates
  one of three vehicles; matching its selection outscores reporting all
  three. \emph{(b)}~Both perceive correctly; near
  clause-for-clause parallelism beats different words.}
  \label{fig:qualitative}
\end{figure}

\subsection{Answer Specialist: 27B QLoRA, Answer-Only SFT}
\label{sec:method-answerer}

The answer specialist is a Qwen~3.6~27B VLM fine-tuned with
QLoRA~\cite{dettmers2023qlora}, and it exists to win the seven open-ended
tasks. Since these are scored against references whose register only the
training answers sample (\cref{sec:task-annotations}), we treat them as
distribution matching rather than reasoning and supervise on the answer
alone: no trace, no schema, no deliberation. With no
reasoning scaffold to lean on, the model must carry the register in its
weights, a capacity question on which the 27B
backbone~\cite{qwen2026qwen36} outperforms the 9B
(\cref{sec:exp-ablation}). Training uses the backbone's non-thinking mode
with the chat template's empty think block masked from the loss
(\cref{sec:exp-impl}), and targets the merged coarse-to-fine Scene-group
answer (\cref{sec:method-tarre}). The result is the ``recite'' half of
\ourmethod{}: the model routinely emits near-paraphrases of references it
has never seen (the splits are video-disjoint; \cref{sec:exp-impl}), and
\cref{fig:qualitative} traces the specialists' open-ended gap to register
and content selection rather than perception.

\paragraph{Where it excels, and where it does not.}
Stronger on the open-ended tasks, the model is weaker on the closed-form
ones, where it asserts while the reasoning specialist verifies: on
validation items whose verdict hinges on inspecting the video, it
recites the salient prior (a collision implies a red light was run) where
the reasoning specialist cites timestamped \texttt{<evidence>} and
returns the correct verdict. BERTScore barely separates such a false
verdict from a true one ($0.57$ vs.\ $0.61$), but the exact-match tasks
punish it outright; this per-question complementarity is what the judge
(\cref{sec:method-judge}) exists to exploit.

\subsection{Judge-Guided Selection}
\label{sec:method-judge}
For every test question (MCQ-OE excepted; \cref{sec:method-overview}),
a frozen, out-of-the-box Qwen~3.6~27B receives the video frames, the
question, and both candidate answers and returns which one to submit:
no fine-tuning and no precomputed routing table. The prompt casts the
judge as a \emph{conservative verifier} rather than a free selector,
a design dictated by its measured position bias and family
self-preference (\cref{sec:exp-ablation}): the candidate expected to
win, the output of the specialist with the higher validation score on
the question's task type, is presented last and is kept unless clearly
contradicted by the video, and a brief per-task instruction states
what separates a correct answer from an incorrect one. A deterministic
pre-filter settles uncontested items, so the judge adjudicates only
where the candidates genuinely disagree.

\begin{table}[!t]
  \caption{Top~6 teams on the AI City Challenge 2026 Track~3 in-domain
  (TAR) leaderboard. BCQ and MCQ are exact-match accuracy; the remaining
  seven tasks (-OE: with explanation; OQA: open-ended QA; T-Desc: temporal
  event description; Summ.: video summarization) are BERTScore F1; Mean is
  the unweighted average over the nine scored task types. Best value per
  column in \textbf{bold}.}
  \label{tab:leaderboard}
  \centering
  \scriptsize
  \setlength{\tabcolsep}{3pt}
  \resizebox{\textwidth}{!}{%
  \begin{tabular}{@{}clcccccccccc@{}}
    \toprule
    \# & Team & Mean & BCQ & MCQ & BCQ-OE & MCQ-OE & OQA & Causal & Scene & T-Desc & Summ. \\
    \midrule
    1 & Stellarview AI (ours) & \textbf{0.6788} & \textbf{1.000} & 0.950 & \textbf{0.669} & 0.969 & 0.499 & \textbf{0.531} & 0.437 & \textbf{0.514} & 0.541 \\
    2 & FPT AI Vision         & 0.6696 & 0.963 & 0.950 & 0.631 & 0.952 & \textbf{0.522} & 0.525 & \textbf{0.445} & 0.498 & \textbf{0.542} \\
    3 & Smart Vision          & 0.6669 & \textbf{1.000} & 0.925 & 0.634 & 0.967 & 0.501 & 0.528 & 0.437 & 0.485 & 0.525 \\
    4 & CHT-IVS               & 0.6563 & 0.988 & 0.938 & 0.643 & 0.957 & 0.465 & 0.516 & 0.406 & 0.491 & 0.503 \\
    5 & CHT\_IVS              & 0.6562 & 0.988 & 0.938 & 0.642 & 0.957 & 0.465 & 0.516 & 0.405 & 0.491 & 0.503 \\
    6 & first\_gogogo         & 0.6369 & 0.994 & \textbf{0.975} & 0.619 & \textbf{0.973} & 0.464 & 0.470 & 0.374 & 0.445 & 0.420 \\
    \bottomrule
  \end{tabular}}
\end{table}

\section{Experiments}
\label{sec:experiments}

\subsection{Implementation Details}
\label{sec:exp-impl}

\paragraph{Hardware and quantization.}
All training runs on a single GPU: the 9B reasoning specialist on one
NVIDIA RTX~5090 (32\,GB), the 27B answer specialist on one NVIDIA
RTX~PRO~6000 Blackwell (96\,GB), with the 27B's frozen base quantized to
4-bit NormalFloat while bf16 LoRA adapters train; the vision tower is
exempt and stays in bf16, since the answer specialist's value depends on
perception.

\paragraph{Training stack.}
At the time of writing, no widely used fine-tuning framework offers stable
support for video inputs to the Qwen~3.5/3.6 series, so we implement our
own training and evaluation pipeline on top of \texttt{transformers}, TRL,
PEFT, and \texttt{bitsandbytes}, keeping prompt construction byte-identical
between training and inference. A chunked loss computes the
cross-entropy in slices, never materializing the full logits tensor over
the large vocabulary; we release the pipeline with our code.

\paragraph{Fine-tuning recipe.}
Both specialists share one recipe. Their backbones are
hybrids~\cite{qwen2026qwen35} interleaving Gated DeltaNet linear
attention~\cite{yang2025gateddeltanet} with full softmax attention in a
fixed 3:1 pattern, with a SwiGLU MLP in every layer; rank-64 LoRA
adapters ($\alpha=128$, dropout $0.05$, no bias) attach to every linear
projection of all three block types: the full-attention $q$/$k$/$v$/$o$
maps, the Gated DeltaNet query/key/value streams with their decay,
update, and output gates, and the SwiGLU gate/up/down maps, twelve
module types in all; the mixer's non-matrix parameters, the vision
tower, and the token embeddings stay frozen. Training uses
maximum sequence length $5120$, a fixed $32$
frames per video at a per-frame budget of $131{,}072$ pixels (floor
$3{,}136$), AdamW with a peak learning rate of $2\times10^{-4}$
on a cosine schedule with $3\%$ warmup and no weight decay, a per-device
batch size of $1$ with gradient accumulation $16$, bf16 precision,
gradient checkpointing, and FlashAttention-2; the 9B trains for four
epochs, the 27B for two. The 27B otherwise differs only as
\cref{sec:method} dictates, training the same adapters over its
NF4-quantized base and keeping them unmerged at inference. All inference,
including the judge's, uses greedy decoding.

\paragraph{Data splits and hygiene.}
The TAR-RE data is split $80/10/10$ into training, validation, and internal
test sets by \emph{video}: all annotations of a video share a split, so no
validation or internal-test video appears in training under any task type,
and the assignment is stratified by source dataset (seed $42$), with
uniform task weighting. Both specialists train on the $80\%$ split; the
validation split drives every development-time decision (checkpoint
selection, the judge's prompt and pre-filter; the judge's weights stay
frozen); the internal test split is never trained on, drives no
development-time decision, and is reserved for the ablations of
\cref{sec:exp-ablation}, whose numbers therefore estimate uncontaminated
generalization.

\subsection{Main Results}
\label{sec:exp-main}

\cref{tab:leaderboard} reports the final in-domain (TAR) leaderboard.
\ourmethod{} ranks \textbf{first} with a mean
score of 0.6788. The win comes from
breadth rather than a single task: our submission attains the best score
on four of the nine scored task types (perfect binary event verification
(BCQ, $100\%$), event verification with explanation, causal linkage, and
temporal event description) and remains within $0.025$ of the per-column
best on every other task. The leaderboard scores the combined submission; the
per-column pattern matches the division of labor seen on validation
(\cref{sec:exp-ablation}).

\subsection{Ablation Study}
\label{sec:exp-ablation}

\begin{table}[!t]
  \caption{Reasoning-supervision ablation on the 9B backbone: identical
  recipe (\cref{sec:exp-impl}), training target either the gold answer
  alone or the full TAR-RE think-then-answer trace
  (\cref{sec:method-reasoner}); greedy decoding on identical items from a
  subset of the internal test split. BCQ and MCQ are exact-match
  accuracy, the rest BERTScore F1; $\Delta$ is reasoning minus
  answer-only.}
  \label{tab:reasoning}
  \centering
  \footnotesize
  \begin{tabular}{@{}lrccr@{}}
    \toprule
    Task & \#Items & Answer-only & TAR-RE reasoning & $\Delta$ \\
    \midrule
    BCQ               & 140 & 0.9571 & 0.9714 & $+0.0143$ \\
    MCQ               &  70 & 0.9429 & 1.0000 & $+0.0571$ \\
    BCQ-OE            & 140 & 0.6972 & 0.6460 & $-0.0512$ \\
    Open QA           &  70 & 0.4426 & 0.4618 & $+0.0192$ \\
    Scene description &  70 & 0.4453 & 0.4620 & $+0.0167$ \\
    \bottomrule
  \end{tabular}
\end{table}

\begin{table}[!t]
  \caption{Ablation of the selection strategy on the internal test split:
  $70$ videos, $490$ items over the six free-text task types (one item per
  video and task; two for BCQ-OE). Mean score is the average over items.}
  \label{tab:selection}
  \centering
  \footnotesize
  \begin{tabular}{@{}lc@{}}
    \toprule
    Selection strategy & Mean score \\
    \midrule
    Reasoning specialist (9B) only      & 0.4460 \\
    Answer specialist (27B) only        & 0.4958 \\
    Judge, neutral rubric               & 0.4887 \\
    Judge-guided selection (ours)       & 0.5043 \\
    \midrule
    Oracle per-question selection       & 0.5347 \\
    \bottomrule
  \end{tabular}
\end{table}

\paragraph{Reasoning supervision.}
\Cref{tab:reasoning} isolates the value of the TAR-RE traces on the 9B
backbone. Trace supervision decides the closed-form tasks: multiple
choice reaches $100\%$ and binary verification also gains. The picture
inverts on BCQ-OE, which the answer-only arm wins by $5.1$ BERTScore
points, while open QA and scene description sit within two points.
Reasoning supervision pays where answers are scored exactly and costs
where similarity to the reference register is scored, the division of
labor the two specialists instantiate.

\paragraph{Selection strategy.}
\Cref{tab:selection} isolates the selection mechanism on the internal
test split. Because the answer specialist wins every free-text task on
average on validation, a static per-task router collapses to the
27B-only row, so $0.4958$ is the ceiling of any routing table.
Judge-guided selection scores $0.5043$; gains over the 27B can arise
only where the judge selected the reasoning specialist's answer, so
this margin is realized per-question flip recovery, not routing. The
prompt is what makes the recovery positive: a neutral ``pick the
better answer'' rubric scores $0.4887$, below the 27B alone,
overturning good defaults on position and verbosity more often than
content earns correct overrides, so the $0.0156$ between the two
rubrics is purely the value of constraining the judge. An oracle
reaches $0.5347$; the judge recovers roughly a fifth of that gap, the
remainder being the price of guarding against its own biases,
quantified below.

\paragraph{Complementarity, not routing.}
The oracle's headroom is located in \emph{flips}: items whose better
answer comes from the specialist that loses that task on average,
precisely the answers static task-to-model
routing~\cite{kyem2025taskspecific,kachhadiya2025multiagent} discards by
construction. On the internal test split, flips are 41\% of BCQ-OE items
(worth $+0.060$ per flipped item under oracle selection), 49\% of causal
linkage ($+0.035$), 37\% of temporal description ($+0.027$), 24\% of
open-ended QA ($+0.024$), and 13\% of scene description ($+0.008$); only
video summarization, at 0\%, would be fully served by a routing table.
The recoverable score concentrates in the reasoning-heavy tasks, hence
arbitration per question rather than per task.

\paragraph{Judge behavior.}
Three measurements shape how the judge is deployed. First, position
bias~\cite{zheng2023judging}: under the neutral rubric the judge favors
the last-presented candidate on 65--79\% of items regardless of author,
and only 51.6\% of its verdicts survive swapping the order (chance
level). A judge this position-driven cannot be granted free choice;
the winner-last ordering of \cref{sec:method-judge} aligns this prior
with the validation-favored candidate, so an override must be earned
by visual evidence rather than by slot. Second,
self-preference~\cite{panickssery2024selfpreference}: averaged over
both orders, the neutral judge selects the same-family 27B candidate on
42.8\% of items against the oracle's 37.8\%. Third, who can judge at all: a 9B
sibling returns the last-presented candidate on every item, so the
frozen 27B is the smallest family member with any content sensitivity,
and neither specialist can take the role, since judging a comparison
containing one's own answer forfeits independence.

\begin{table}[!t]
  \caption{Levers that did not work. Deltas are BERTScore~F1 against the
  base configuration on the validation data (leaderboard where marked LB);
  paired bootstrap puts the noise floor at roughly $0.012$--$0.016$, so
  smaller deltas are ties.}
  \label{tab:negative}
  \centering
  \scriptsize
  \setlength{\tabcolsep}{4pt}
  \resizebox{\textwidth}{!}{%
  \begin{tabular}{@{}llrl@{}}
    \toprule
    Task & Lever & $\Delta$ & Failure mode \\
    \midrule
    \multicolumn{4}{@{}l}{\emph{Rewriting the answer after generation}} \\
    OQA    & Text-only LLM rewrites the answers      & $-0.071$ & polishes away gold-matching detail \\
    BCQ-OE & LLM cleanup of drafts                   & $-0.012$ & deletes words that match the gold \\
    BCQ-OE & LLM fusion of four candidates           & $-0.016$ & invents detail the drafts agree is absent \\
    Scene  & Trim last sentence / keep first three   & $-0.028$ / $-0.074$ & the metric is recall-driven \\
    \multicolumn{4}{@{}l}{\emph{Injecting content from outside the model}} \\
    Scene  & Vehicle-enumeration prompting           & $-0.014$ & recall $+0.04$, precision $-0.07$ \\
    Scene  & Appending gold open-QA content          & $-0.083$ & event narration is not scene inventory \\
    OQA    & Sibling-description hints in the prompt & $-0.016$ & propagates the hints' errors \\
    \multicolumn{4}{@{}l}{\emph{Adding capacity or input resolution}} \\
    Scene  & Vision-tower LoRA (last 4 / 8 blocks)   & $+0.002$ / $+0.003$ & rotates hallucinations, does not reduce them \\
    Scene  & rsLoRA, rank 128                        & $-0.012$ & shorter outputs drop marginal attributes \\
    Scene  & $32\!\to\!64$ frames; larger pixel budget & $\sim 0$ & input levers saturated \\
    Scene  & Few frames at very high resolution      & $-0.077$ (LB) & transcribes distant incidental signage \\
    \bottomrule
  \end{tabular}}
\end{table}

\subsection{What Did Not Work}
\label{sec:exp-negative}

\cref{tab:negative} lists our main negative results; three lessons stand out.

\paragraph{A single reference cannot be paraphrased toward.}
Every post-hoc transform that reworded a generated answer regressed: LLM
rewriting, cleanup, fusion, and every truncation. The metric rewards
token-level proximity to one specific reference, and a paraphrase made
without that reference in hand moves away from it on average; proximity
must instead be trained into the weights: hence the answer
specialist's design, and why the judge selects rather than edits. The
only edits that survived injected verifiable content extracted with
Gemini 3.1 Pro and left surrounding text untouched: transcribing the
location overlay burned into each test frame (read from a
native-resolution crop) into the scene descriptions ($+0.004$ LB), and
anchoring open-ended answers to event windows Gemini localized from
cropped clips, checked against sibling-task facts ($+0.008$ LB).

\paragraph{The remaining headroom points to perception, not capacity.}
On scene description, grafting in the references' vehicle sentences by
oracle gains $+0.06$, so headroom exists. But every deployable route to
it traded a small recall gain for a larger precision loss, and every
capacity lever (vision-tower LoRA, higher rank, more frames, more
pixels) tied or regressed: the binding constraint is the precision of
what the model asserts it saw, and only fine-tuning improved it.

\paragraph{Local validation is not a leaderboard proxy.}
Our best-ever validation configuration for scene description scored
$0.077$ \emph{below} the shipped model on the leaderboard. The two sets
are drawn differently: the organizer test set is dominated by short,
collision-heavy clips (median length $\sim$10\,s), whereas our local
split follows the released-data distribution. A configuration tuned on
the latter (few frames, very high resolution) reads and transcribes
storefront signage in the sign-dense test domain, collapsing precision.
After this inversion we trusted validation only for the component it
directly measures and verified configurations on the evaluation
server, as permitted.

\section{Conclusion}
\label{sec:conclusion}

We presented \ourmethod{}, the winning entry to the Traffic Anomaly
Reasoning track of the 2026 AI City Challenge: an annotation audit,
reasoning supervision regenerated into a structured, timestamp-grounded
schema (TAR-RE), two complementary specialists trained on it, and a
training-free LLM judge arbitrating per question, ranking first with
mean score 0.6788.
Our analysis shows unconstrained multimodal judging is unreliable;
position-bias control is what makes arbitration profitable.
The broader lesson: when a benchmark's supervision is auto-labeled and
its metric scores similarity to a single reference, reasoning quality
and reference matching are separable objectives, and conflating them in
one model leaves accuracy on the table.

\bibliographystyle{splncs04}
\bibliography{references}

\end{document}